\documentclass[11pt]{article}

\usepackage[final]{acl}
\usepackage{times}
\usepackage{latexsym}
\usepackage{amsmath} 
\usepackage{csvsimple}  
\usepackage[table]{xcolor}

\usepackage{booktabs}
\usepackage{geometry}
\usepackage{amsthm}
\usepackage{stfloats}
\usepackage{caption}

\usepackage{xcolor}
\definecolor{msftBlack}{RGB}{0,0,0}

\usepackage[T1]{fontenc}
\usepackage[utf8]{inputenc}

\usepackage{microtype}

\usepackage{inconsolata}

\usepackage{graphicx}

\usepackage[shortlabels,inline]{enumitem}
\usepackage{tikz,lipsum}
\usepackage[most]{tcolorbox}
\usepackage{ragged2e}
\usepackage[dvipsnames]{xcolor}
\usepackage{amsmath}
\usepackage{booktabs}
\usepackage{tabularray}
\usepackage{arydshln}
\usepackage{stmaryrd}
\usepackage{marvosym}
\usepackage{colortbl}
\usepackage{multicol}
\usepackage{multirow}
\usepackage{float}

\usepackage{cleveref}
\crefname{section}{\S}{\S}
\crefname{table}{Table}{Tables}
\crefname{figure}{Fig.}{Figs.}
\crefname{algorithm}{Alg.}{}
\crefname{ALC@unique}{Line}{Lines}
\crefname{equation}{Eq.}{Eqs.}
\crefname{appendix}{App.}{Apps.}
\crefformat{section}{\S#2#1#3} 

\usepackage{soul}
\definecolor{tablegray}{RGB}{223, 242, 252}

\usepackage{todonotes}

\NewDocumentCommand{\prompt}{O{} +m}{%
\begin{tcolorbox}[
    coltitle=white,
    colframe=black,
    colback=black!5!white,
    boxrule=1pt,
    enhanced jigsaw,
    breakable,
    pad at break*=2mm,
    left=2pt,
    right=2pt,
    top=2pt,
    bottom=2pt,
    fontupper=\small,
    fontlower=\small,
    title={#1}, % Optional title
]
#2 % Multiline text content
\end{tcolorbox}
}

\usepackage[tikz]{bclogo}

\title{Exploring Cross-Lingual Knowledge Transfer via Transliteration-Based MLM Fine-Tuning for Critically Low-resource Chakma Language}
\author{
Adity Khisa\\
IIT, University of Dhaka \\
\texttt{bsse1334@iit.du.ac.bd}
\And
Nusrat Jahan Lia \\
IIT, University of Dhaka \\
\texttt{bsse1306@iit.du.ac.bd}
\And
Tasnim Mahfuz Nafis \\
IIT, University of Dhaka \\
\texttt{bsse1327@iit.du.ac.bd}
\AND
Zarif Masud \\
Toronto Metropolitan University \\
\texttt{zarif.masud@gmail.com}
\And
Tanzir Pial\\
Stony Brook University\\
\texttt{tpial@cs.stonybrook.edu}
\AND
Shebuti Rayana \\
State University of New York at Old Westbury\\
\texttt{rayanas@oldwestbury.edu}
\And
Ahmedul Kabir \\
IIT, University of Dhaka \\
\texttt{kabir@iit.du.ac.bd}
}
\begin{document}
\maketitle
%\todo{Use capitalization for Section 1, Table 1. looks better imo.}
\begin{abstract}
As an Indo-Aryan language with limited available data, Chakma remains largely underrepresented in language models. In this work, we introduce a novel corpus of contextually coherent Bangla-transliterated Chakma, curated from Chakma literature, and validated by native speakers. Using this dataset, we fine-tune six encoder-based transformer models, including multilingual (mBERT, XLM-RoBERTa, DistilBERT), regional (BanglaBERT, IndicBERT), and monolingual English (DeBERTaV3) variants on masked language modeling (MLM) tasks. Our experiments show that fine-tuned multilingual models outperform their pre-trained counterparts when adapted to Bangla-transliterated Chakma, achieving up to 73.54\% token accuracy and a perplexity as low as 2.90. Our analysis further highlights the impact of data quality on model performance and shows the limitations of OCR pipelines for morphologically rich Indic scripts. Our research demonstrates that Bangla-transliterated Chakma can be very effective for transfer learning for Chakma language, and we release our dataset\footnote{\url{https://github.com/adity1234567/Chakma-MLM-Dataset.git}} to encourage further research on multilingual language modeling for low-resource languages. 

\end{abstract}

\section{Introduction}

Large Language Models (LLMs) have transformed the Natural Language Processing (NLP) world through unsupervised pre-training using large corpora of unlabeled data. Since labeled data are not required, LLMs can take advantage of the huge text corpora available in the public domain. For example, even first-generation language models such as BERT use a corpus of 3.3 billion English words \citep{devlin2019bert}, while more recent LLMs use multiple massive corpora such as RedPajama \citep{weber2024redpajama} scraped from the web with hundreds of trillions of tokens. 
%\vspace{\baselineskip}
\begin{figure*}[t]  
\centering
\includegraphics[width=\textwidth]{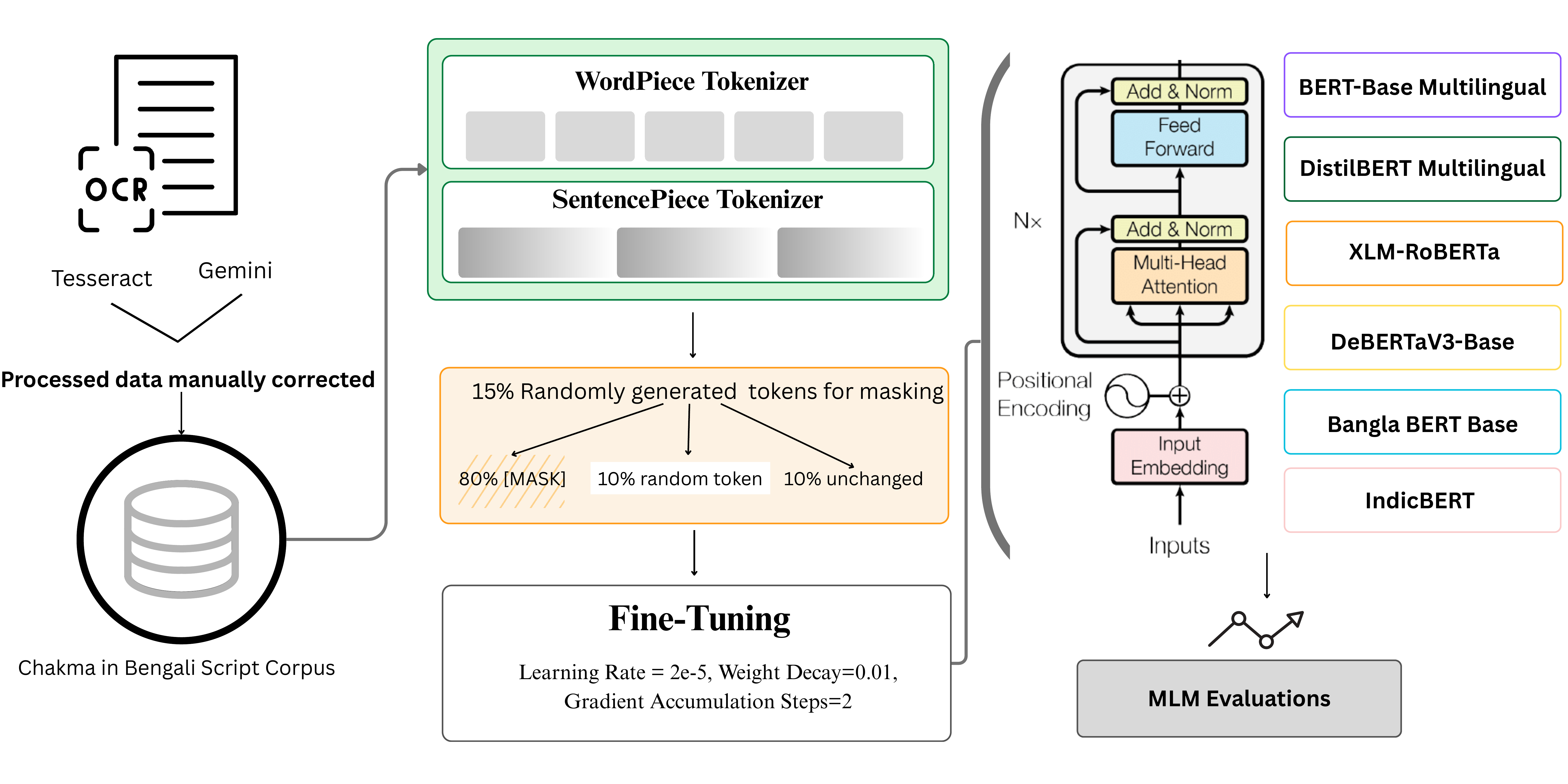}  
\caption{Overall workflow of OCR-based data curation, manual correction, and MLM fine-tuning for Bangla-transliterated Chakma language model}
\label{fig:methodology}
\end{figure*}
However, the sheer volume of data required for pre-training LLMs poses a challenge for low-resource languages even without labels, as seen in some recent works using datasets of 15 million words for Māori \citep{james2022code_switch}, 332 million tokens for Swahili \citep{conneau2020xlmr}, and 108 million tokens from 11 African languages for AfriBERTa \citep{ogueji2021smalldata}. 
Compared to the trillions of tokens available in high-resource languages, these million-scale corpora are minuscule. Consequently, training LLMs with low-resource corpora does not yield good results, as upon encountering new vocabulary, expressions, or culturally specific semantics, the models struggle to utilize their training patterns for accurate understanding and generation \cite{zhong2024opportunities}.
  
To address this limitation, researchers have explored knowledge transfer, from LLMs trained on high-resource languages through Masked Language Model (MLM) fine-tuning on the comparatively lower resource language corpus \citep{fernando2025masking}. %\todo{I would like a citation for this sentence.} 
\citet{muller2020being} further showed that we can leverage the same transfer learning benefit through transliteration when the two languages do not share a script. They achieved a significant performance gain for Uyghur (105K sentences) and Sorani Kurdish (380K sentences) transliterated into the Latin script, compared to pre-training on those data in their original script alone. %\vspace{\baselineskip}

Chakma is an Indo-Aryan language, used as a first language by roughly one million people from the Chakma community living across parts of Bangladesh, India and Myanmar \citep{cadc2025chakma}. Although Chakma has its own script Ojhā Pāṭh, a considerable portion of Chakma literature is produced in Bangla transliteration \citep{brandt2018writing}. Chakma remains a low-resource language with data scarcity both in its original script and Bangla transliteration \citep{chakma2024chakmanmt}. At the same time, Bangla script is regularly used in training of multilingual LLMs like mBERT \citep{pires2019multilingual}. In this context, our work shows that Bangla-transliterated Chakma dataset can yield moderately strong performance through MLM fine-tuning. Following \citet{muller2020being}'s idea for transliteration, we use Chakma text transliterated in Bangla for MLM-tuning multiple LLMs that are pre-trained on Bangla. Since, to the best of our knowledge, no contextually coherent Bangla-transliterated Chakma corpus exists, we have curated a novel corpus from Chakma books containing 4,570 manually validated sentences to run our experiments.

Our major contributions are as follows:
\begin{itemize}
    \item We develop a novel Bangla-transliterated Chakma dataset, curated from images sourced from four books of Chakma literature using Tesseract OCR, comprising a total number of 6,353 sentences, of which 4,570 have been manually corrected. 
    \item We show that language models can learn low-resource languages via MLM fine-tuning on the script of a related language, as demonstrated using Bangla script to fine-tune a Chakma model. 
    \item We demonstrate how data quality impacts model performance, showing that better OCR for Bangla script, compatible with Bangla-transliterated Chakma, can significantly improve transfer learning for the Chakma language.
\end{itemize}

\section{Related Works}
%\todo{Need to add a one or two line intro}
In this section, we discuss four key areas that inform our work: multilingual NLP, model adaptation and quantization, tokenization and morphological challenges in Indic scripts, and existing language resources for Bangla and Chakma. These topics collectively highlight the progress and challenges in building effective models for low-resource languages like Chakma.

\subsection{Multilingual NLP}
The evolution of multilingual models has been driven by the need to extend transformer-based models to low-resource languages, particularly those with limited
data or non-Latin scripts \cite{pakray2025natural}. \citet{devlin2019bert} introduced BERT along with its multilingual variant mBERT, and this marked a turning point. Models like mBERT, pretrained on Wikipedia data across 104 languages using WordPiece (vocabulary of 110K tokens), and XLM-R,
trained on CommonCrawl data from 100 languages with a 250K SentencePiece vocabulary \cite{conneau2019unsupervised}, enabled zero-shot cross-lingual transfer. XLM-R, relying solely on MLM pretraining, achieved state-of-the-art performance on multiple benchmarks \cite{ebrahimi2021adapt}.
%\vspace{\baselineskip}

These multilingual models often show strong zero-shot performance, but disparities remain: languages with less pretraining data or non-Latin scripts typically lag behind high-resource languages \cite{ebrahimi2021adapt,marchisio2024does}. For example, \citet{wu2020all} and \citet{muller2020being} show that mBERT’s zero-shot accuracy varies widely by language, with some “hard” languages (often low-resource or using different scripts) remaining poorly served without additional adaptation. These
findings spurred research leveraging pretrained transformer models and specialized techniques to handle underrepresented languages \cite{tela2020transferring,hangya2022improving,bharadiya2023transfer,pakray2025natural}. Our work builds on this by fine-tuning a Chakma-specific MLM encoder, addressing data scarcity for this low-resource Indic language.

\subsection{Model adaptation techniques and quantization}
To address performance disparities in low-resource languages, adaptation strategies emerged to tailor pre-trained models to specific languages or domains. When more data are available, continued monolingual pre-training in target-language data, as demonstrated by \citet{chau2020parsing}, improved zero-shot performance, while domain-adaptive MLM pre-training improves downstream performance even in low-resource settings \cite{gururangan2020don}. Another strategy is to expand the vocabulary of a multilingual model to better cover the target language’s lexicon and then additional MLM training  improves performance for underrepresented languages \citep{wang2020extending}. 

\subsection{Tokenization and morphology in Indic scripts}

Indic languages like Bangla and Chakma are morphologically rich and use complex abugida scripts \cite{chowdhury2025phonological}, which raise challenges for subword tokenization. Standard BPE or WordPiece tokenizers can fragment important morphological units, hurting model performance \cite{pattnayak2025tokenization}. Recent work demonstrates that SentencePiece (unigram) tokenization often preserves morphological information better than BPE for Indic languages. For instance, \citet{pattnayak2025tokenization} found that, for zero-shot named entity recognition across several Indic languages, a SentencePiece-based vocabulary outperformed BPE, because it more cleanly segments root words and affixes. Others have noted that vowel forms in abugida scripts (\textit{matras}) attach to consonants and can appear above, below, or beside the base character, which makes character-level segmentation non-trivial \cite{kashid2024roundtripocr,maung2025hybrid}. 

\subsection{Bangla and Chakma language resources}
Although \citet{joshi2020state} categorize Bangla among languages lacking labeled data,  \citet{bhattacharjee2021banglabert} developed BanglaBERT, a BERT-base model on the \textit{Bangla2B+} corpus with  2.18 billion tokens from Bangla text, and introduced the Bangla Language Understanding Benchmark (BLUB). BanglaBERT achieves state-of-the-art results on multiple Bangla NLU tasks, outperforming both multilingual baselines (mBERT, XLM-R) and previous monolingual models.  

\begin{figure*}[t!]
    \centering
    \includegraphics[width=\textwidth]{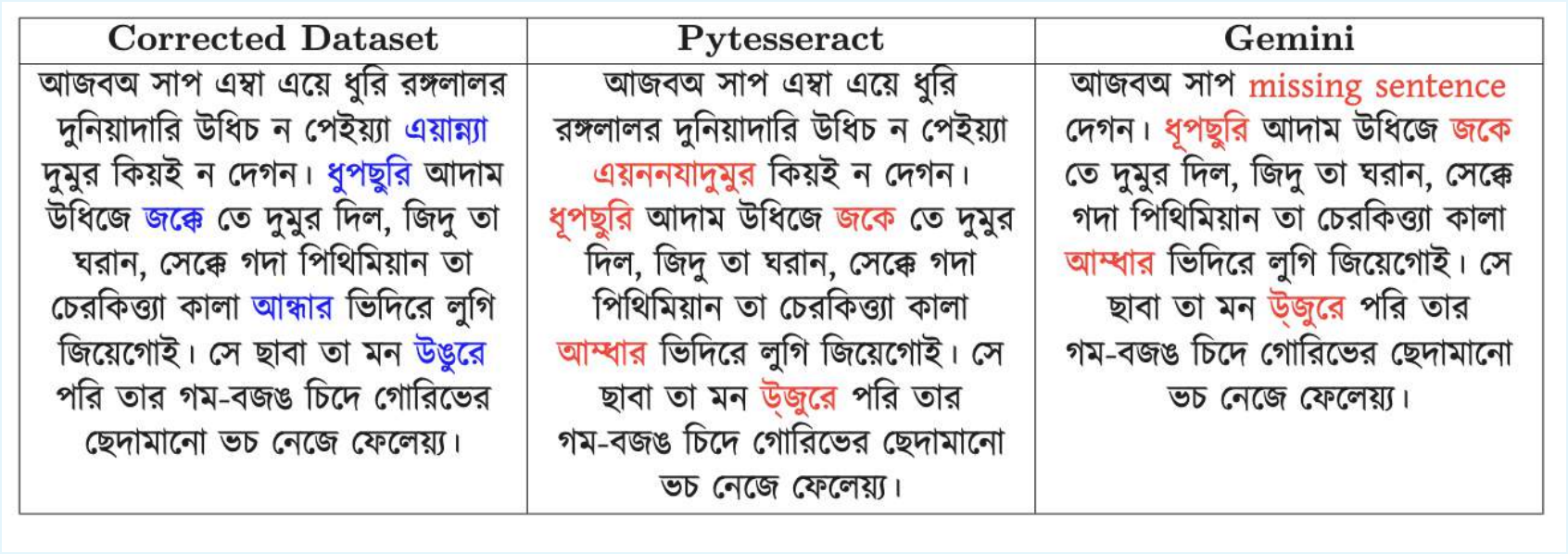}
    \caption{Sample data illustrating quality comparison across different methods, highlighting \textbf{missing sentences} in Gemini and \textbf{spelling errors} in other models caused by the misinterpretation of conjunct characters, phonetic signs, vowel diacritics, consonant modifiers, nasalization, and related orthographic features.}
    \label{fig:ocr-comparison}
\end{figure*}

\renewcommand{\arraystretch}{1.3} % Increase row height
\begin{table*}[ht]
\scriptsize
\resizebox{\textwidth}{!}{
\begin{tabular}{|l|l|l|l|}
\hline
\textbf{Model} & \textbf{Vocab Size} & \textbf{Tokenizer \& Special Tokens} & \textbf{Tokenization Method} \\ \hline
\textbf{BERT-Base Multilingual (cased)} & $\sim$120k (WordPiece) & {[CLS] ... [SEP]}, [MASK] & WordPiece \\ \hline
\textbf{DistilBERT Multilingual (cased)} & $\sim$120k (WordPiece) & {[CLS] ... [SEP]}, [MASK] & WordPiece \\ \hline
\textbf{XLM-RoBERTa (XLM-R)} & $\sim$250k (SentencePiece) & $\langle s \rangle$ ... $\langle/ s \rangle$, $\langle mask \rangle$ & SentencePiece \\ \hline
\textbf{DeBERTaV3-Base} & $\sim$128k (WordPiece-style) & {[CLS] ... [SEP]}, [MASK] & WordPiece-style \\ \hline
\textbf{Bangla BERT Base} & $\sim$32k (WordPiece) & {[CLS] ... [SEP]}, [MASK] & WordPiece \\ \hline
\textbf{IndicBERT} & $\sim$200k (SentencePiece) & $\langle s \rangle$ ... $\langle/ s \rangle$, $\langle mask \rangle$ & SentencePiece \\ \hline
\end{tabular}
}
\caption{Comparison of encoder-based models used in our evaluation. Differences arise in vocabulary size, tokenizer conventions, and tokenization methods. The models include BERT-Base Multilingual \cite{devlin2019bert}, DistilBERT Multilingual \cite{sanh2019distilbert}, XLM-RoBERTa \cite{liu2019roberta}, DeBERTaV3-Base \cite{he2021debertav3}, Bangla BERT \cite{Sagor_2020}, and IndicBERT \cite{kakwani2020indicnlpsuite}.}
\label{tab:model-details}
\end{table*}

In contrast, NLP work on Chakma is scant. The first known work in Chakma NLP effort is ChakmaNMT \cite{chakma2024chakmanmt}, which constructed the first parallel corpus (15K sentence-pairs translation, from Chakma to Bangla) and trained a translation model. Using BanglaT5 and transliteration-based back-translation, they achieved a BLEU score of 17.8 for Chakma to Bangla translation. However, this work does not include the Bangla-transliterated Chakma text. The MELD dataset \cite{mahi2025meld} compiled transliterated sentence-level text in Chakma (and Garo, Marma) using the Bangla script. We opted not to use MELD, as its collection of isolated sentences lacks the semantic coherence required for our study. Instead, we focus on Bangla-transliterated Chakma texts extracted via OCR from printed literature.

\section{Dataset Creation}

Emphasizing the \textbf{authenticity} of linguistic resources, particularly in the field where the digitized materials are scarce and under-resourced, we construct a novel dataset combining four books (novels and poems) written in the Chakma language, utilizing Bangla script (see Figure~\ref{fig:book-metadata} and Table~\ref{tab:ocr-stats}). To collect these materials, we directly engaged with scholars whose first language is Chakma. The books were gathered from libraries on the basis of their recommendations. However, we acknowledge that most scholars prioritize the preservation and use of their own Chakma script. \citet{chakma2024chakmanmt} also assigns importance to the Chakma script. Most pre-trained models lack support for the complex structure of Chakma scripts. 
The Bangla-transliterated Chakma script enables the models to process the language effectively using their existing tokenizers.

\subsection{Corpus Compilation: Sources and Scale}

The data were extracted from the image of the pages of the books using \textbf{three primary methods}: Pytesseract \cite{pytesseract}, Gemini \cite{comanici2025gemini}, and manual processing.
We used the PyTesseract OCR model and the Gemini 2.5 Pro model API separately to independently assess the quality of text extraction from different systems.

%% the whole process
 PyTesseract encountered problems with the recognition of Bangla's conjunctive characters and committed frequent spelling errors, as presented in Figure~\ref{fig:ocr-comparison}.  On the other hand, Gemini 2.5 Pro with the free API, posed usage restrictions creating a barrier to scalability as we processed 400 images. Moreover, the Gemini API deviated from correctness in alphabet recognition, often produced incomplete sentences, and sometimes omitted entire sentences, affecting the overall quality of the extracted text. Some examples are presented in Figure~\ref{fig:ocr-comparison}.

 Due to these limitations, we manually fixed one book entirely and another book partially, which we discuss in Section~\ref{subsec:manual}. The dataset is split into training, testing and validation subsets, as shown in Table~\ref{tab:ocr-stats}.

\begin{table}[t]
\centering
\small 
\begin{tabular}{l l}
\toprule
\textbf{Dataset} & \shortstack{\textbf{Dataset Split} \\ \textbf{(sentences)}} \\
\midrule
\textbf{Tesseract OCR (Tes OCR)} &
\begin{tabular}[t]{@{}l@{}}
train: 4,348 \\
eval: 832 \\
test: 1,173 
\end{tabular} \\
\addlinespace
\textbf{Gemini OCR} &
\begin{tabular}[t]{@{}l@{}}
train: 3,815 \\
eval: 994 \\
test: 1,173 
\end{tabular} \\
\addlinespace
\textbf{Manually Fixed Data} &
\begin{tabular}[t]{@{}l@{}}
train: 2,908 \\
eval: 545 \\
test: 1,118 
\end{tabular} \\
\bottomrule
\end{tabular}
\caption{Breakdown of training, evaluation, and test sentence counts for datasets obtained from Tesseract OCR, Gemini OCR outputs, and manually corrected data.}
\label{tab:ocr-stats}
\end{table}
\begin{table*}[ht]
%\centering

\scriptsize
\renewcommand{\arraystretch}{1.8}
\resizebox{\textwidth}{!}{
\begin{tabular}{|l|ccc|ccc|}
\hline
\multirow{2}{*}{\textbf{Model}} 
& \multicolumn{3}{c|}{\textbf{Accuracy (\%) ↑}} 
& \multicolumn{3}{c|}{\textbf{Perplexity ↓}} \\ \cline{2-7}
& \shortstack[c]{\rule{0pt}{2.4ex}Without \\ MLM}

& \shortstack[c]{ With \\ MLM} 
& \textbf{Performance} 
& \shortstack[c]{ Without \\ MLM} 
& \shortstack[c]{ With \\ MLM} 
& \textbf{Performance} \\ \hline

DeBERTaV3-Base          & 0.00  & 72.08 & +72.08  & 39329757.5 &\textbf{2.90}
&-39329754.6 \\ \hline
XLM-RoBERTa             & 46.24 & \textbf{73.54}
 & +27.30  & 24.39 & 3.27 & -21.12  \\ \hline
BERT-Base mBERT         & \textbf{48.43} & 70.00 & +21.57  & \textbf{13.12}  & 4.017 & -9.103  \\ \hline
DistilBERT Multilingual & 38.78 & 65.08 & +26.30  & 24.284 & 4.3046 & -19.978  \\ \hline
Bangla BERT Base        & 29.87 & 54.52 & +24.65  & 250.09 & 11.79 & -238.3  \\ \hline
IndicBERT               & 17.54 & 45.36 & +27.82  & 1823.61 & 16.79 & -1806.82 \\ \hline
\end{tabular}
}
\caption{Performance comparison of models before and after MLM fine-tuning using manually annotated Chakma corpora. Accuracy (\%) and perplexity are reported. Lower perplexity indicates better language modeling performance.} 
%\todo{Instead of highlighting with red, just make them bold.}}
\label{tab:mlm-tuning-impact}
\end{table*}

\subsection{Manual Curation for Linguistic Fidelity} \label{subsec:manual}
Both the OCR models and LLMs struggled with accurate processing of conjunct
characters, phonetic signs, including vowel diacritics, consonant modifiers, nasalization, silent consonant and incompleteness of sentences (Figure~\ref{fig:ocr-comparison}). These complex character clusters are fundamental to the Bangla orthography but often are misinterpreted or omitted by OCR systems due to their non-linear composition and script variability  \cite{ali-etal-2023-gold,guo-etal-2023-pipeline}. After identifying the limitations and to ensure the linguistic fidelity of our dataset, 4,570 sentences of OCR and LLM extracted text underwent a multi-stage manual correction and validation process. 
Two co-authors of this paper rectified these specific errors to guarantee the high integrity and usability of the final manual dataset. The overall workflow of the paper is presented in Figure~\ref{fig:methodology}.

\section{MLM-tuning for low-resource languages}

We fine-tuned six encoder-based models (including monolingual, multilingual and regional variants) on limited Chakma text written in Bangla script and compared their performance. Table~\ref{tab:model-details} summarizes the models used in our experiments, including their vocabulary sizes and tokenization algorithms. All of these models have been pre-trained on Bangla before.

These LLMS are trained to predict the probability of a masked token/word given the context of surrounding words. This gives the models a foundational understanding of trained languages that can be generalized to other tasks \citep{wolf-etal-2020-transformers}. Although each model comes in a similar-sized 12-layer configuration (270M–300M parameters for base models),  they vary in vocabulary sizes, tokenizer types, special tokens, and critically, the dataset they were first pre-trained on. 

%\subsection{Model and Training Data Setup }
 For MLM fine-tuning, we masked 15\% of tokens in each input sequence using the standard masking strategy: 80\% replaced with the appropriate mask token ([MASK] or <mask>), 10\% substituted with random vocabulary tokens, and 10\% left unchanged. We maintained strict separation between training, validation, and test datasets across all experiments to prevent data leakage.

After multiple trials and errors, in our final configuration, we use the Adam optimizer, with a learning rate of $2 \times 10^{-5}$ for all the models. The maximum number of epochs is 20. The dropout rate is 0.01. We keep the batch size at 8.
For testing the fine-tuned  models, we ensure consistency and reproducibility across the models.

\section{Results}
We evaluated the performance across three different data processing pipelines (Pytesseract, Gemini and manual processing) and also compared both universal and regional model types. Following the works of \citet{salazar2019masked}, \citet{rogers2021primer} and  \citet{ethayarajh2019contextual}, we evaluate our MLM fine-tuned models using perplexity, masked token accuracy, precision, recall, F1(macro), pseudo-log-likelihood (PLL) and predictive entropy.

\subsection{Language Modeling Capability}
\textit{RQ1}: How effective are the pre-trained language models at masked language modeling for the (monolingual) Chakma language written in the Bangla script? \\

Table~\ref{tab:mlm-tuning-impact} shows that fine-tuned encoder-based language models consistently outperform their pre-trained counterparts for Bangla-transliterated Chakma. The fine-tuned models achieve accuracies up to 73.54\% (XLM-RoBERTa) and perplexity as low as 2.899 (DeBERTaV3-Base) on manually corrected data, underscoring the value of adaptation for low-resource languages. Notably, monolingual models like DeBERTaV3-Base, which start with no prior knowledge of Chakma or Bangla script (0\% baseline accuracy), achieve competitive results post-fine-tuning, demonstrating the robustness of adaptation even without cross-lingual pre-training. MLM-tuning also yields a marked reduction in prediction entropy, indicating increased confidence in masked-token predictions in Table~\ref{tab:model-comparison}.

\begin{figure}[t]
    %\centering
    \includegraphics[width=\linewidth]{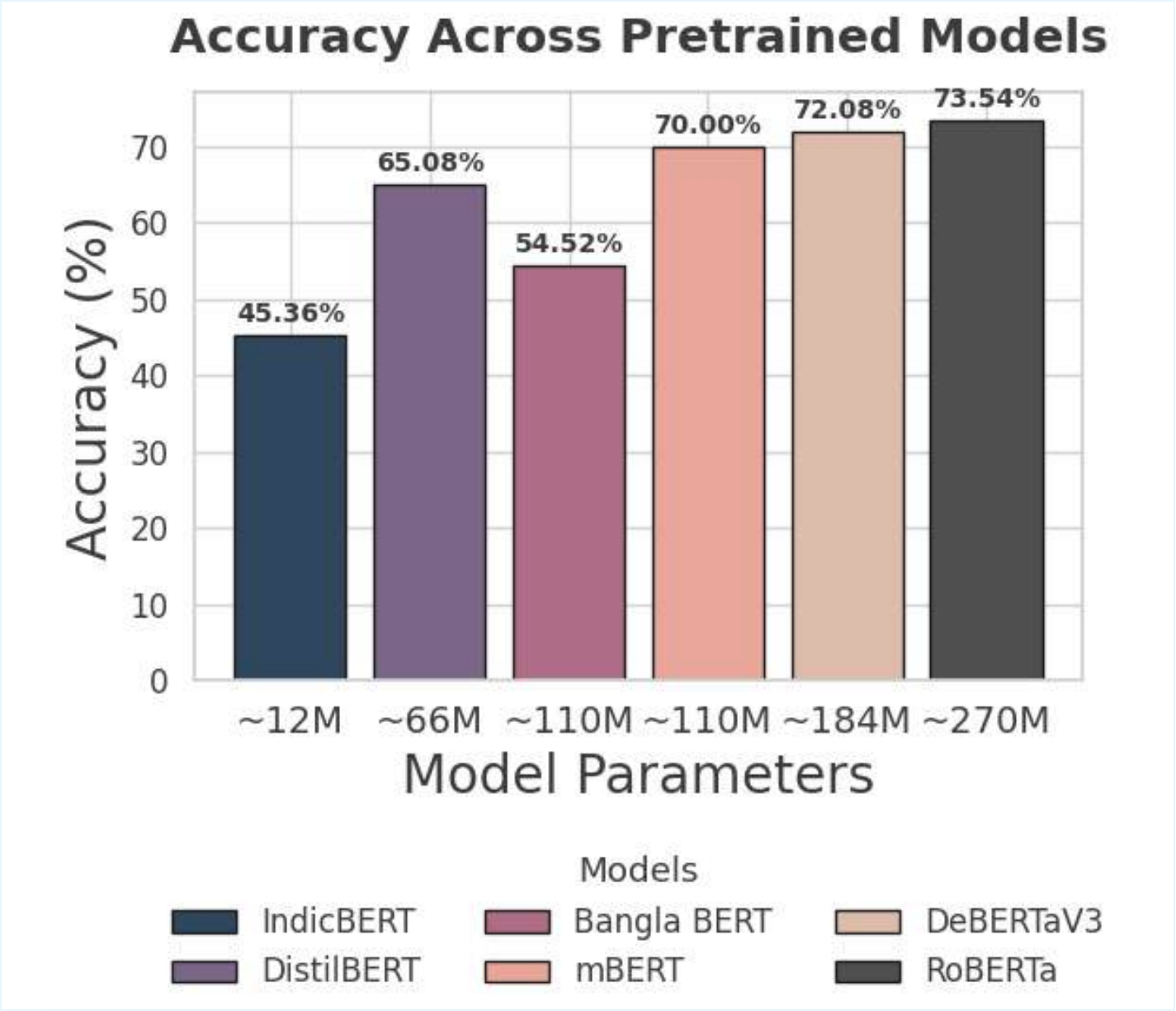}
    \caption{Comparison of universal multilingual and regional encoder models. Each grouped bar chart is showing the accuracy of pre-trained language models fine-tuned on manually fixed data, categorized by their parameter sizes.}
    \label{fig:accuracy_token}
\end{figure}

Our best perplexity score of 2.899 is substantially lower (indicating better performance) than the perplexity scores reported for BERT on English datasets~\cite{salazar2019masked}. We treat this as an empirical observation rather than definitive evidence of superior absolute performance. This low perplexity may be an artifact of our dataset characteristics, including its relatively small size and the specific nature of the data (potentially featuring simpler or more repetitive linguistic structures compared to diverse English corpora). Hypotheses for this include reduced lexical diversity or script-specific tokenization efficiencies in Chakma, but the exact reasons remain unclear and could be explored in more detail in future work, perhaps by evaluating on larger, more varied Chakma datasets.
\paragraph{Outperforming of universal models over regional encoders:}
From Figure~\ref{fig:accuracy_token} and Table~\ref{tab:data-quality-impact}, we observe a consistent advantage for multilingual encoder models (XLM-RoBERTa, BERT-Base mBERT, DistilBERT Multilingual) and the monolingual DeBERTaV3-Base over regional encoder models (BanglaBERT, IndicBERT). 
Because tokenizers and vocabulary sizes differ across models, masked-language accuracy and perplexity are computed on model-specific tokenizations rather than an identical token sequence. This can potentially advantage models that produce fewer tokens per input, since they evaluate fewer positions and may face fewer rare-subword predictions. However, we argue that the comparison remains informative: the vocabularies are not extremely different, and the underlying dataset is identical for all models, and that accuracy is not simply determined by token count (see Table~\ref{tab:model-details} and Table~\ref{tab:mlm-tuning-impact}). 

We analyze two primary factors that influence model effectiveness: model parameter size and tokenization efficiency.

\textbf{1. Parameter size → tokenization robustness.} Larger multilingual models are trained on broader, more diverse corpora and typically learn richer subword vocabularies. This reduces out-of-vocabulary occurrences, over-fragmentation, and tokenization drift, which can otherwise harm downstream performance. These effects can cause some tokenizers to produce 2–3 times more tokens for the same input (see Figure~\ref{fig:accuracy_token}) \cite{rust2020good}.

\textbf{2. Tokenizer efficiency → evaluation metrics.} A smaller number of tokens allows each token to carry more semantic context and reduces prediction noise for masked positions. Over-fragmentation, by contrast, spreads probability mass across many rare subwords, penalizing sequence-level scoring and hurting pseudo-log-likelihood (PLL) \cite{kudo2018sentencepiece}.

 \subsection{Impact of Data Quality}
%\subsection{Evaluation Metrics}
\begin{table*}[ht]
\centering
\scriptsize
\renewcommand{\arraystretch}{1.6}
\setlength{\tabcolsep}{3.2pt}

\resizebox{\textwidth}{!}{%
\begin{tabular}{|p{2.8cm}|cc|cc|cc|cc|cc|}
\hline
\multirow{3}{*}{\textbf{Model}} 
& \multicolumn{2}{c|}{\textbf{Manual}} 
& \multicolumn{4}{c|}{\textbf{Tesseract}} 
& \multicolumn{4}{c|}{\textbf{Gemini}} \\ \cline{2-11}

&  &
& \multicolumn{2}{c|}{\textbf{Self-Finetuned}} 
& \multicolumn{2}{c|}{\textbf{Manual-Finetuned}} 
& \multicolumn{2}{c|}{\textbf{Self-Finetuned}} 
& \multicolumn{2}{c|}{\textbf{Manual-Finetuned}} \\ \cline{2-11}

& \textbf{Acc.(↑)} & \textbf{PPL(↓)} 
& \textbf{Acc.(↑)} & \textbf{PPL(↓)} 
& \textbf{Acc.(↑)} & \textbf{PPL(↓)} 
& \textbf{Acc.(↑)} & \textbf{PPL(↓)} 
& \textbf{Acc.(↑)} & \textbf{PPL(↓)} 
\\ \hline

DeBERTaV3-Base         
& 72.08 & 2.90  
& 46.94 & 7.82 & 46.52 & 9.75 & 46.86 & 9.12 & 45.77 & 10.01 \\ \hline

XLM-RoBERTa          
& 73.54 & 3.27  
& 30.28 & 54.39 & 29.14 & 77.74 & 28.70 & 80.16 & 29.67 & 78.04 \\ \hline

BERT-Base mBERT     
& 70.00 & 4.02  
& 32.91 & 27.50 & 31.56 & 40.59 & 31.75 & 44.16 & 31.23 & 42.35 \\ \hline

DistilBERT Multilingual
& 65.08 & 4.30  
& 31.64 & 29.05 & 29.63 & 43.91 & 30.49 & 43.20 & 30.04 & 41.96 \\ \hline

Bangla BERT Base     
& 54.52 & 11.79 
& 22.17 & 299.54 & 20.19 & 384.28 & 20.71 & 483.57 & 20.86 & 467.25 \\ \hline

IndicBERT            
& 45.36 & 16.79 
& 23.04 & 83.67 & 24.05 & 80.18 & 23.64 & 97.62 & 23.12 & 103.10 \\ \hline

\end{tabular}
} % end of resizebox

\caption{Impact of data quality on model performance. Accuracy (\%) and Perplexity (PPL) are reported for each model fine-tuned on manually annotated, Tesseract-processed, and Gemini-processed data. In our table, \textit{Self-Finetuned} refers to training and evaluating each model on the same dataset, while \textit{Manual-Finetuned} involves training on manually corrected data but evaluating on other test datasets like Tesseract or Gemini test sets.}
\label{tab:data-quality-impact}
\end{table*}
%\textit{RQ2:} What is the impact of data source quality (manual, OCR, LLM-extracted) on MLM performance for Chakma?\\

\textit{RQ2:} In the context of the morphologically rich Bangla-transliterated Chakma, how does the OCR noise of data affect MLM performance?

Building on the findings from RQ1, where fine-tuning encoder-based models on manually corrected Chakma data demonstrated strong improvements in masked language modeling capabilities, we now explore the extent to which OCR-induced noise (stemming from script-specific challenges like transliteration variations and complex conjunct consonants) disrupts the learning of morphological structures in Bangla transliterated Chakma. In each case, the models were fine-tuned and evaluated on their respective dataset, which we refer to as \textit{Self-Finetuned} in our Table~\ref{tab:data-quality-impact}. Additionally, we evaluated the model fine-tuned on the manually corrected dataset against the Tesseract and Gemini 2.5 Pro test sets, which we denote as \textit{Manually-Finetuned} in the Table~\ref{tab:data-quality-impact}.
Due to transliteration-induced variation with more complex conjunct consonants, the transliterated data (Bangla-transliterated Chakma) appears morphologically heavier than Bangla.

From the Table~\ref{tab:data-quality-impact}, we can see that models trained with Tesseract and Gemini 2.5 Pro processed data struggled to grasp the Chakma language, showing limited improvements even after fine-tuning, particularly evident in cases where models like DeBERTaV3-Base\cite{he2021debertav3} had no initial understanding of Chakma (Table~\ref{tab:mlm-tuning-impact}). The fine-tuning with the manually fixed dataset led to substantial gains in accuracy, highlighting that the model learns the affixes, inflections and complex forms of the language in a better way. Meanwhile, these models drop their performance when testing on the noisy test dataset. For instance, the XLM-RoBERTa model achieved its strongest performance with manual data, far surpassing its baseline and revealing that noisy OCR outputs can actually degrade model capabilities compared to their pre-fine-tuned state.
 
We find a similar pattern when examining perplexity across datasets for individual models. From Table~\ref{tab:data-quality-impact}, the manual dataset consistently yielded low perplexity, indicating strong language modeling and coherence. However, Tesseract and Gemini data introduced higher perplexity, often worsening it beyond the base model's levels due to inherent noise and errors. This trend holds across all six models in our experiments, emphasizing how high-quality data refines predictions while OCR-generated inaccuracies amplify confusion. Furthermore, when testing manually fine-tuned models on Tesseract or Gemini data, their perplexity suffered slightly compared to self-fine-tuned counterparts, reinforcing the pervasive impact of noise in OCR pipelines on overall model robustness.

Overall, these results show the critical role of preserving morphologically accurate data quality in enhancing model performance for low-resource indigenous languages like Chakma.

\section{Conclusion}

In this work, we introduced a Bangla-transliterated Chakma dataset, derived from Chakma literature using Tesseract, Gemini 2.5 Pro OCR and manual transcription. We empirically demonstrate that pre-trained multilingual language models can be effectively adapted for the Chakma language through fine-tuning on this data, establishing a strong baseline for Masked Language Modeling for Chakma.
Our comprehensive experiments further underscore that model performance is highly sensitive to data quality, and that iterative cleaning directly enhances model performance. To support future research, we publicly release our manually refined dataset. 
A compelling direction for future work is to investigate the optimal transliteration target for low-resource languages. We hypothesize that for Chakma, which shares significant typological and lexical similarity with Bangla, transliteration into the Bangla script may yield superior performance compared to the English script, despite the generally stronger pre-training of LLMs on English. Systematically evaluating this trade-off between linguistic proximity and model capability remains an open question.

\section{ Limitations and Future Work}

This study focuses on understanding the potential of LLM adaptability to low-resource languages. In our work, we have considered Chakma language as a case study.
However, our manually validated Bangla-transliterated Chakma language dataset contains only 4570 sentences. The sentences are collected from story books, which is not sufficient to reflect diverse real-world scenarios, especially in a modern context. So, we aim to expand our Chakma corpus incorporating more diverse text sources, including spoken language transcripts, community-generated contents and parallel translations. 
Transliteration of Chakma dataset to Latin script is another direction of research following the works of \citet{muller2020being}. If such a dataset exists, we can test the hypothesis that transliterating Chakma to a related language (Bangla) as opposed to the strongest language (English) may yield better performance.
Inspired by \citet{devlin2019bert}, we can test our fine-tuned model for Next Sentence Prediction (NSP) accuracy to get a better understanding of how well our model is understanding the Chakma language. 
Improving OCR accuracy to extract the text with a better performance for conjunct characters, phonetic signs including vowel diacritics, consonant modifiers, nasalization, and others is also a potential direction for improvement.

\section{Acknowledgement}
 The authors would like to thank Bijoy Chakma (Sociology, University of Rajshahi) and Shriswa Chakma (Department of Management Studies, Bangladesh University of Professionals) for their contributions to data collection and manual processing. 
\bibliography{custom}
\clearpage
\appendix
%\onecolumn

\section{Appendix}
\label{sec:appendix}
\begin{table}[ht]
% \centering
\scriptsize
\caption{Model Fine-tuned on Manual Data: Cross-Test Performance}
\label{tab:finetuned-manual}
\renewcommand{\arraystretch}{1.3}
\resizebox{\textwidth}{!}{%
\begin{tabular}{|c|c|c|c|c|c|c|c|c|c|c|c|}
\hline
\textbf{Model} & \textbf{Data} & \textbf{Loss} & \textbf{Perplexity} & \textbf{\shortstack{Accuracy(\%)\\(masked token acc.)}} & \textbf{Precision} & \textbf{Recall} & \textbf{F1\_macro} & \textbf{\shortstack{Prediction\\Entropy}} & \textbf{\shortstack{Pseudo-log\\likelihood}} & \textbf{\shortstack{Evaluated \\Tokens (sentences)}} \\
\hline

\multirow{3}{*}{\shortstack{manual\_roberta}} 
  & manual             & 3.1942     &  24.3903       & 46.24     & 0.3432    & 0.2799     & 0.2876     & 2.8258     & -3.1910       & 8092 (431)\\
\cline{2-11}
  & teserrect          & 4.3534  & 77.7443  & 29.14    & 0.2502 & 0.1469 & 0.1603 & 2.7436 & -4.3550  & 10713 (569) \\
\cline{2-11}
  & gemini-2.5-pro     & 4.3573  & 78.0479  & 29.67    & 0.253  & 0.1576 & 0.1677 & 2.7591 & -4.3559  & 10645 (569) \\
\hline

\multirow{3}{*}{manual\_bert}
  & manual             & 1.3905  & 4.017    & 70    & 0.5107 & 0.4353 & 0.4512 & 1.0178 & -1.3916  & 9011 (495) \\
\cline{2-11}
  & teserrect          & 3.7036  & 40.5924  & 31.56    & 0.2886 & 0.1837 & 0.1974 & 2.123  & -3.7023  & 12198 (657) \\
\cline{2-11}
  & gemini-2.5-pro     & 3.7461  & 42.3548  & 31.23    & 0.2738 & 0.1716 & 0.1880 & 2.1382 & -3.7426  & 12445 (657) \\
\hline

\multirow{3}{*}{manual\_distilbert}
  & manual             & 1.4740  & 4.3668   & 67.10    & 0.4803 & 0.4074 & 0.4074 & 1.2122 & -1.4742  & 9180 (495) \\
\cline{2-11}
  & teserrect          & 3.7822  & 43.9145  & 29.63    & 0.233  & 0.1415 & 0.1521 & 2.3025 & -3.7787  & 12516 (657) \\
\cline{2-11}
  & gemini-2.5-pro     & 3.7368  & 41.9629  & 30.04    & 0.2357 & 0.1568 & 0.1636 & 2.3025 & -3.7411  & 12424 (657) \\
\hline

\multirow{3}{*}{manual\_debarta}
  & manual             & 1.0644  & 2.8991   & 72.08    & 0.452  & 0.3549 & 0.3724 & 0.8393 & -1.0654  & 16202 (864) \\
\cline{2-11}
  & teserrect          & 2.2775  & 9.7526   & 46.52    & 0.1911 & 0.1402 & 0.1446 & 1.3831 & -2.2782  & 20466 (1085) \\
\cline{2-11}
  & gemini-2.5-pro     & 2.3037  & 10.0116  & 45.77    & 0.1999 & 0.1460 & 0.1515 & 1.3957 & -2.3066  & 20402 (1085) \\
\hline

\multirow{3}{*}{sagorbangla}
  & manual             & 2.4675  & 11.7925  & 54.52 & 0.367  & 0.2663 & 0.2861 & 2.6335 & -2.4707  & 18587 (986) \\
\cline{2-11}
  & teserrect          & 6.1659  & 476.2059 & 20.45    & 0.1542 & 0.0979 & 0.1042 & 3.9245 & -6.1638  & 8216 (434) \\
\cline{2-11}
  & gemini-2.5-pro     & 6.2530  & 519.5888 & 20.51    & 0.1343 & 0.0826 & 0.0886 & 3.9889 & -6.2558  & 8148 (434) \\
\hline

\multirow{3}{*}{IndicBERT}
  & manual             & 2.4492     & 11.5791       & 52.75      &  0.4337     &  0.2994     & 0.3297     & 2.4188     & -2.4546       & 6627 (351) \\
\cline{2-11}
  & teserrect          & 4.6562      & 105.2307       & 23.03      & 0.2572     &  0.1140     & 0.1385     & 3.7807     & -4.6625      & 8392 (429)\\
\cline{2-11}
  & gemini-2.5-pro     & 4.5815      & 97.6632       & 23.19      & 0.2964     & 0.1357    & 0.1649     & 3.7785     & -4.5871       & 8251 (429) \\
\hline

\end{tabular}

}
\end{table}

\begin{table}[H]
% \centering
\scriptsize
\caption{Model Performance Comparison %\todo{Let's break up this table into three tables/figures. One highlighting the data quality impact, the second highlighting the model impact. Finally, the last and the most important one highlighting mlm-tuning impact.}
%\todo{ It's good to have all of these metrics in an appendix. But in the main section, it is distracting the reader from realizing what they are supposed to learn from the table. Use only one or at most two metrics in the tables.}
}
\label{tab:model-comparison}
\resizebox{\textwidth}{!}{%
\begin{tabular}{|c|c|c|c|c|c|c|c|c|c|c|c|}
\hline
\textbf{Model} & \textbf{Data} & \textbf{Type} & \textbf{Loss} & \textbf{Perplexity} & \textbf{\shortstack{Accuracy\\(masked token accuracy)}} & \textbf{Precision} & \textbf{Recall} & \textbf{F1\_macro} & \textbf{\shortstack{Prediction\\Entropy}} & \textbf{\shortstack{Pseudo-log-\\likelihood}} & \textbf{Tokens} \\
\hline

% BERT-Base Multilingual (cased)
\multirow{6}{*}{\centering\raisebox{-0.5\height}{\shortstack{BERT-Base\\Multilingual (cased)}}}
& \multirow{2}{*}{\centering\raisebox{-0.5\height}{manual}} 
  & base      & 2.5744  & 13.1234 & 48.43\% & 0.4173 & 0.3144 & 0.3321 & 2.3547 & -2.5763 & 9126 (495) \\
\cline{3-12}
&                            & finetuned & 1.3905  & 4.017   & 70\%    & 0.5107 & 0.4353 & 0.4512 & 1.0178 & -1.3916 & 9011 (495) \\
\cline{2-12}
& \multirow{2}{*}{\centering\raisebox{-0.5\height}{teserrect}} 
  & base      & 3.8187  & 45.5459 & 25.05\% & 0.2491 & 0.1507 & 0.1640 & 3.3034 & -3.8194 & 12371 (657) \\
\cline{3-12}
&                            & finetuned & 3.3143  & 27.5039 & 32.91\% & 0.2882 & 0.1706 & 0.1889 & 2.6401 & -3.3211 & 12493 (657) \\
\cline{2-12}
& \multirow{2}{*}{\centering\raisebox{-0.5\height}{gemini-2.5-pro}} 
  & base      & 3.7713  & 43.4369 & 26.19\% & 0.2492 & 0.1543 & 0.1695 & 3.2742 & -3.7731 & 12338 (657) \\
\cline{3-12}
&                            & finetuned & 3.7878  & 44.1608 & 31.75\% & 0.2391 & 0.1821 & 0.1855 & 2.1291 & -3.7898 & 12353 (657) \\
\hline

% DistilBERT Multilingual (cased)
\multirow{6}{*}{\centering\raisebox{-0.5\height}{\shortstack{DistilBERT\\Multilingual (cased)}}}
& \multirow{2}{*}{\centering\raisebox{-0.5\height}{manual}} 
  & base      & 3.1898  & 24.284  & 38.78\% & 0.3030 & 0.2044 & 0.2135 & 3.2877 & -3.1989 & 9106 (495) \\
\cline{3-12}
&                            & finetuned & 1.4597  & 4.3046  & 65.08\% & 0.5070 & 0.4045 & 0.4323 & 1.4180 & -1.4636 & 8796 (439) \\
\cline{2-12}
& \multirow{2}{*}{\centering\raisebox{-0.5\height}{teserrect}} 
  & base      & 4.2519  & 70.2371 & 20.10\% & 0.2064 & 0.1030 & 0.1090 & 3.8775 & -4.2527 & 12126 (657) \\
\cline{3-12}
&                            & finetuned & 3.3691  & 29.052  & 31.64\% & 0.2845 & 0.1699 & 0.1872 & 2.6530 & -3.3700 & 1197 (657) \\
\cline{2-12}
& \multirow{2}{*}{\centering\raisebox{-0.5\height}{gemini-2.5-pro}} 
  & base      & 4.2025  & 66.8516 & 20.30\% & 0.2175 & 0.1087 & 0.1150 & 3.8539 & -4.2038 & 12459 (657) \\
\cline{3-12}
&                            & finetuned & 3.7658  & 43.197  & 30.49\% & 0.2533 & 0.1701 & 0.1811 & 2.3147 & -3.7675 & 12353 (657) \\
\hline

% XLM-RoBERTa (XLM-R)
\multirow{6}{*}{\centering\raisebox{-0.5\height}{\shortstack{XLM-RoBERTa\\(XLM-R)}}}
& \multirow{2}{*}{\centering\raisebox{-0.5\height}{manual}} 
  & base      & 3.1942  & 24.3903 & 46.24\% & 0.3432 & 0.2799 & 0.2876 & 2.8258 & -3.1910 & 8092 (431) \\
\cline{3-12}
&                            & finetuned & 1.1858  & 3.2732  & 73.54\% & 0.5677 & 0.5114 & 0.5219 & 1.0693 & -1.1918 & 8043 (431) \\
\cline{2-12}
& \multirow{2}{*}{\centering\raisebox{-0.5\height}{teserrect}} 
  & base      & 4.5293  & 92.6939 & 25.79\% & 0.2294 & 0.1498 & 0.1619 & 4.0378 & -4.5321 & 10653 (569) \\
\cline{3-12}
&                            & finetuned & 3.14      & 54.39   & 30.28\% & 0.2634 & 0.1554 & 0.1719 & 3.2284 & -3.9981 & 10327 (569) \\
\cline{2-12}
& \multirow{2}{*}{\centering\raisebox{-0.5\height}{gemini-2.5-pro}} 
  & base      & 4.5354  & 93.2627 & 25.72\% & 0.2213 & 0.1417 & 0.1552 & 4.0543 & -4.5400 & 10762 (569) \\
\cline{3-12}
&                            & finetuned & 4.3840  & 80.1618 & 28.70\% & 0.2421 & 0.1575 & 0.1631 & 2.8692 & -4.3910 & 10620 (569) \\
\hline

% DeBERTaV3-Base
\multirow{6}{*}{\centering\raisebox{-0.5\height}{\shortstack{DeBERTaV3\\Base}}}
& \multirow{2}{*}{\centering\raisebox{-0.5\height}{manual}} 
  & base      & 17.4875 & 39329757.5 & 0\%     & 0.0000 & 0.0000 & 0.0000 & 7.8197 & -17.4848 & 16457 (864) \\
\cline{3-12}
&                            & finetuned & 1.0644  & 2.8991  & 72.08\%    & 0.4520 & 0.3549 & 0.3724 & 0.8393 & -1.0654 & 16202 (864) \\
\cline{2-12}
& \multirow{2}{*}{\centering\raisebox{-0.5\height}{teserrect}} 
  & base      & 14.6398 & 2280263.447 & 0\%     & 0.0000 & 0.0000 & 0.0000 & 7.8999 & -14.6415 & 20447 (1085) \\
\cline{3-12}
&                            & finetuned &  2.0572   &  7.8241     & 46.94\%    & 0.3072     & 0.1697    & 0.1859     & 1.562    & -2.0610     & 20477 (1085) \\
\cline{2-12}
& \multirow{2}{*}{\centering\raisebox{-0.5\height}{gemini-2.5-pro}} 
  & base      & 16.1899 & 10744904.22 & 0\%     & 0.0000 & 0.0000 & 0.0000 & 8.1224 & -16.1899 & 20398 (1085) \\
\cline{3-12}
&                            & finetuned & 2.2105    &  9.1200     & 46.86\%     & 0.2412     & 0.1576     & 0.1688     & 1.4322     & -2.2130      & 20509 (1085) \\
\hline

% Bangla BERT Base
\multirow{6}{*}{\centering\raisebox{-0.5\height}{\shortstack{Bangla BERT\\Base}}}
& \multirow{2}{*}{\centering\raisebox{-0.5\height}{manual}} 
  & base      & 5.5218  & 250.0904 & 29.87\% & 0.2338 & 0.1478 & 0.1637 & 5.7929 & -5.5302 & 6451 (344) \\
\cline{3-12}
&                            & finetuned & 2.4675  & 11.7925 & 54.52\% & 0.3670 & 0.2663 & 0.2861 & 2.6335 & -2.4707 & 18587 (986) \\
\cline{2-12}
& \multirow{2}{*}{\centering\raisebox{-0.5\height}{teserrect}} 
  & base      & 6.8747  & 967.5296 & 15.67\% & 0.1516 & 0.0860 & 0.0981 & 6.6578 & -6.8720 & 8129 (434) \\
\cline{3-12}
&                            & finetuned & 5.7022  & 299.5393 & 22.17\% & 0.1598 & 0.1039 & 0.1116 & 4.2600 & -5.7041 & 8167 (434) \\
\cline{2-12}
& \multirow{2}{*}{\centering\raisebox{-0.5\height}{gemini-2.5-pro}} 
  & base      & 6.8133  & 909.9072 & 16.95\% & 0.1514 & 0.0847 & 0.0967 & 6.5669 & -6.8131 & 8256 (434) \\
\cline{3-12}
&                            & finetuned & 6.1469      & 467.2527      & 20.86\%     & 0.1442     &  0.0888     & 0.0953     & 3.9319     & -6.1402      & 8179 (434) \\
\hline

% IndicBERT
\multirow{6}{*}{\centering\raisebox{-0.5\height}{IndicBERT}}
& \multirow{2}{*}{\centering\raisebox{-0.5\height}{manual}} 
  & base      & 7.5091  & 1824.6141 & 17.54\% & 0.3636 & 0.1131 & 0.1489 & 5.3301 & -7.5190 & 18403 (971) \\
\cline{3-12}
&                            & finetuned & 2.8209  & 16.7924  & 45.36\% & 0.4261 & 0.2239 & 0.2693 & 2.8646 & -2.8273 & 18418 (971) \\
\cline{2-12}
& \multirow{2}{*}{\centering\raisebox{-0.5\height}{teserrect}} 
  & base      & 8.0048  & 2995.1897 & 12.62\% & 0.2703 & 0.0822 & 0.1060 & 5.4509 & -8.0154 & 8065 (429) \\
\cline{3-12}
&                            & finetuned & 4.4269  & 83.6677  & 23.04\% & 0.2744 & 0.1141 & 0.1434 & 3.9951 & -4.4268 & 8046 (429) \\
\cline{2-12}
& \multirow{2}{*}{\centering\raisebox{-0.5\height}{gemini-2.5-pro}} 
  & base      & 8.0774  & 3220.7776 & 12.42\% & 0.2762 & 0.0784 & 0.1026 & 5.4241 & -8.0825 & 8130 (429) \\
\cline{3-12}
&                            & finetuned & 4.5811  & 97.6174  & 23.64\% & 0.2733 & 0.1186 & 0.1470 & 3.7354 & -4.5813 & 8046 (429) \\
\hline
\end{tabular}
}
\end{table}

\begin{figure*}[!t]
    \caption{Details of Chakma Storybooks Used in the Dataset}
    \includegraphics[width=\textwidth]{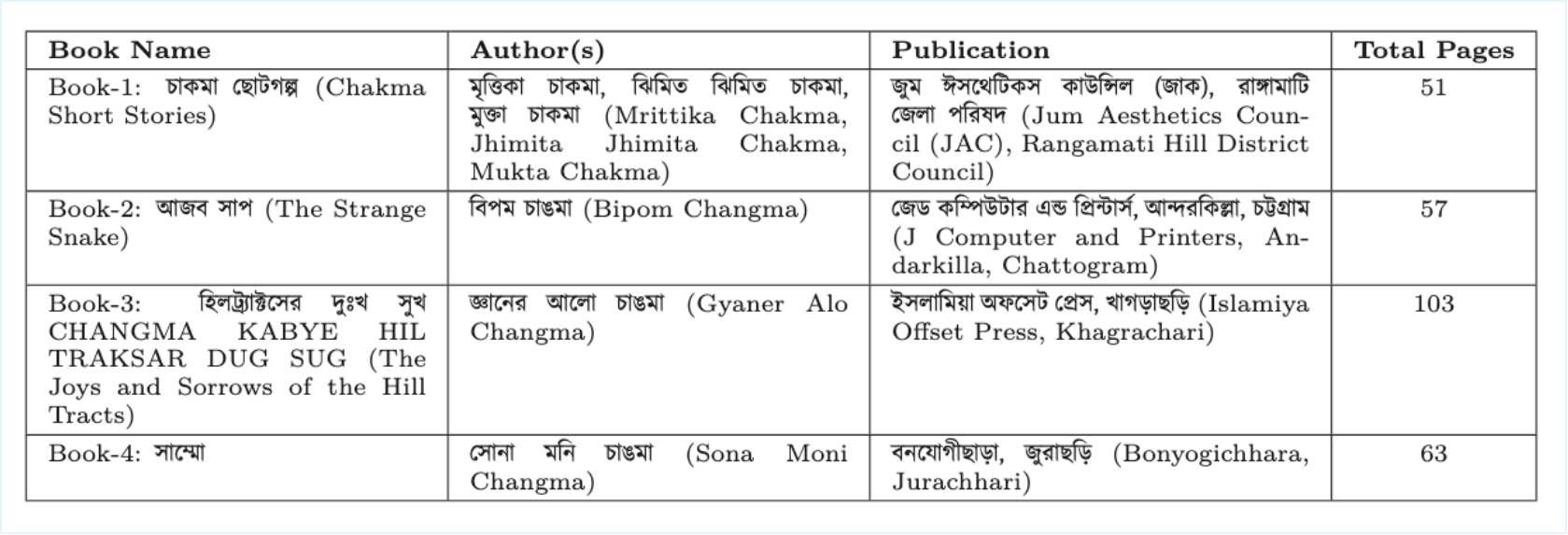}
    \label{fig:book-metadata}
\end{figure*}

\end{document}